\documentclass[sigconf,nonacm]{acmart}
\AtBeginDocument{%
  }

\usepackage[table]{xcolor}
\usepackage{multirow}
\usepackage{graphicx}

\begin{document}

\title{MCPO: Modality-Contrastive Preference Optimization for Multimodal Chain-of-Thought Compression}

\author{Guangheng Yang}
\authornote{Equal contribution}
\email{ygh25@mails.tsinghua.edu.cn}
\affiliation{%
 \institution{Tsinghua University}
  \institution{Huawei Technologies Ltd.}
  \city{Beijing}
  \country{China}
}

\author{Zhenliang Ni}
\authornotemark[1]
\email{nizhenliang2@huawei.com}
\affiliation{%
  \institution{Huawei Technologies Ltd.}
  \city{Beijing}
  \country{China}
}

\author{Zhenkai Wu}
\email{wuzhenkai2@huawei.com}
\affiliation{%
  \institution{Huawei Technologies Ltd.}
  \city{Beijing}
  \country{China}
}

\author{Han Shu}
\authornote{Corresponding authors}
\email{han.shu@huawei.com}
\affiliation{%
  \institution{Huawei Technologies Ltd.}
  \city{Beijing}
  \country{China}
}

\author{Juan Feng}
\authornotemark[2]
\email{juan.feng@sz.tsinghua.edu.cn}
\affiliation{%
  \institution{Tsinghua University}
  \city{Beijing}
  \country{China}
}

\author{Wenming Yang}
\email{yang.wenming@sz.tsinghua.edu.cn}
\affiliation{%
  \institution{Tsinghua University}
  \city{Beijing}
  \country{China}
}

\author{Jie Hu}
\email{hujie23@huawei.com}
\affiliation{%
  \institution{Huawei Technologies Ltd.}
  \city{Shanghai}
  \country{China}
}

\renewcommand{\shortauthors}{Yang et al.}

\begin{abstract}
Recently, multimodal large-scale reasoning models have demonstrated remarkable capabilities in solving complex tasks through long Chains-of-Thought (M-CoT). However, excessively long reasoning trajectories incur substantial computational costs and significant KV-cache pressure. Existing CoT compression and alignment paradigms mainly rely on static rules or single-dimensional preferences, lacking fine-grained cross-modal constraints; as a result, they are prone to inducing visual laziness and hallucinatory reasoning. To address these issues, we propose Modality-Contrastive Preference Optimization (MCPO), a highly sample-efficient two-stage length-compression method that requires fewer than 900 training samples. In the compression stage, we introduce a step-level Normalized Cross-Modal Mutual Information (NCMI) pruning algorithm, which automatically identifies and removes visual-independent reasoning steps by comparing the reasoning discrepancies between with-image and no-image contexts. This significantly reduces redundancy and hallucinatory content in the reasoning chains. In the alignment stage, the model first undergoes supervised fine-tuning to achieve domain-adaptive initialization, followed by optimization using an asymmetric multimodal length-controlled preference loss. This objective adopts a highly nonlinear odds-ratio formulation that provides steep gradients in the with-image context to reinforce length constraints for preferred trajectories, while applying a scaled, flat-gradient linear difference in the no-image context to maintain modality consistency, thereby achieving stable cross-modal preference alignment. Extensive experiments on mainstream base models such as Qwen3-VL-Thinking show that our method can reduce CoT length by up to 69.5\% and achieve up to 3.34$\times$ end-to-end inference speedup while preserving original accuracy.
\end{abstract}

\maketitle

\begin{figure*}[htbp]
  \centering
  \includegraphics[width=\textwidth]{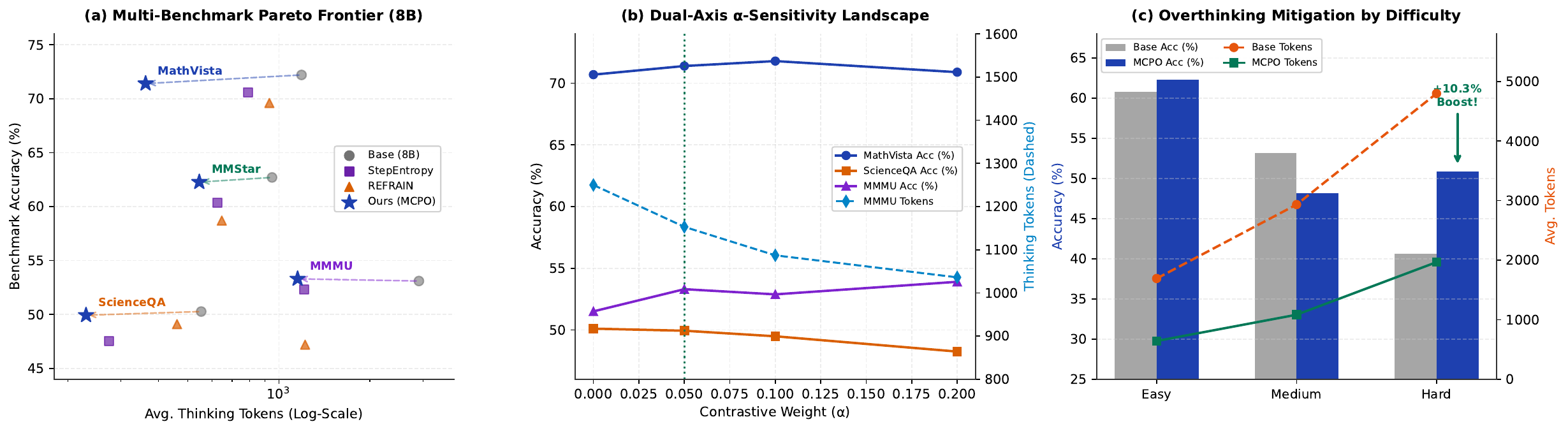}
  \caption{Comprehensive empirical evaluation of the MCPO framework on Qwen3-VL-8B-Thinking. 
  (a) \textbf{Multi-benchmark vector shift} illustrating the optimization trajectory from Base-8B to MCPO across MathVista, ScienceQA, MMMU, and MMStar: thinking token lengths are significantly compressed (by 42.6\%--69.5\%) while maintaining competitive task accuracy compared to StepEntropy and REFRAIN baselines. 
  (b) \textbf{Dual-axis hyperparameter landscape} measuring accuracy and token length sensitivity across contrastive weights $\alpha \in \{0.0, 0.05, 0.1, 0.2\}$, identifying $\alpha=0.05$ as the optimal Pareto choice. 
  (c) \textbf{Overthinking mitigation across reasoning difficulty levels}, demonstrating that MCPO adaptively curtails redundant reasoning steps on simple tasks while unlocking a +10.3\% absolute accuracy boost on Hard reasoning tasks.}
  \label{fig:main_master_8b}
\end{figure*}

\section{Introduction}
In recent years, multimodal large‑scale reasoning models~\cite{guo2025deepseek,team2025qwq,bai2025qwen3} have achieved remarkable progress in tasks such as visual question answering, scene understanding, and cross‑modal reasoning. Multimodal Chain‑of‑Thought (M‑CoT)~\cite{wei2022chain} has emerged as a key mechanism for enhancing complex reasoning capabilities, enabling models to generate multi‑step reasoning trajectories that more precisely interpret visual content, construct cross‑modal associations, and exhibit human‑like reasoning depth in challenging scenarios~\cite{cheng2025objectives,xue2024decompose}. However, these long reasoning chains impose substantial system burdens: inference latency increases significantly, KV‑cache pressure grows sharply, and the overall computational cost becomes prohibitive in real‑world deployment~\cite{feng2025efficient,qu2025survey,liu2025efficient,xu2025scalable,liao2025reward}. As model sizes continue to expand, achieving effective M‑CoT length compression while preserving reasoning quality has become a central challenge for practical multimodal reasoning systems.

To reduce inference overhead, prior work has explored rule‑based pruning, and preference alignment to compress chain‑of‑thought reasoning~\cite{xu2025scalable,muennighoff2025s1,arora2025training,hou2025thinkprune}. Yet these approaches exhibit notable limitations. Many rely on static heuristics or single‑dimensional preferences and lack fine‑grained modeling of visual information, often inducing “visual laziness,” where models ignore image content during reasoning. The absence of cross‑modal consistency constraints further increases the likelihood of hallucinated reasoning, particularly in tasks with strong visual dependence. Moreover, existing methods frequently require large amounts of high‑quality annotated data or costly online preference optimization, resulting in high training overhead and limited applicability in low‑resource settings~\cite{yu2025dapo,chen2025not}. Consequently, current CoT compression paradigms struggle to simultaneously achieve high compression ratios, cross‑modal consistency, and low training cost.

To address these limitations, we propose Modality‑Contrastive Preference Optimization (MCPO), a highly sample‑efficient two‑stage framework for multimodal chain‑of‑thought compression that requires fewer than 900 offline training samples. In the compression stage, MCPO introduces a step‑level Normalized Cross‑Modal Mutual Information (NCMI) pruning algorithm that compares the model’s reasoning behavior under with‑image and no‑image contexts to automatically determine whether each reasoning step genuinely depends on visual information. Steps with low cross‑modal mutual information are pruned, effectively removing visual independent reasoning. This mechanism substantially reduces redundant reasoning and suppresses hallucinations at their source, producing more compact reasoning trajectories that maintain faithful reliance on visual content.

In the alignment stage, MCPO further stabilizes the compressed reasoning style through an asymmetric multimodal length‑controlled preference loss. The model first undergoes supervised fine‑tuning for domain‑adaptive initialization, followed by preference optimization using a novel asymmetric objective. Under with‑image contexts, the loss employs a highly nonlinear odds‑ratio formulation that yields steep gradients, strongly reinforcing the preference for shorter, high‑quality reasoning trajectories. Under no‑image contexts, the loss adopts a scaled linear difference with flatter gradients to preserve cross‑modal consistency and prevent over‑reliance on linguistic patterns. This asymmetric design enables MCPO to maintain visual grounding while reliably controlling reasoning length. Extensive experiments on prominent base models such as Qwen3-VL-Thinking demonstrate that MCPO reduces chain-of-thought length by up to 69.5\% without sacrificing accuracy. This indicates that redundant reasoning steps can be pruned effectively while fully preserving the model's core visual comprehension capabilities. This reduction translates to significant end-to-end execution speedups (up to 3.34$\times$), substantially lowering inference latency and delivering strong efficiency and robustness under extremely low-resource data and compute conditions.
In summary, the main contributions of this work are as follows:
\vspace{-4pt} %
\begin{itemize}
    \setlength{\itemsep}{2pt}  %
    \setlength{\parsep}{0pt}
    \setlength{\parskip}{0pt}
    \item \textbf{Step-Level Cross-Modal Pruning}: We propose an adaptive step-level NCMI pruning method to quantitatively identify and remove visual-independent redundant and hallucinatory reasoning steps prior to training.
    \item \textbf{Asymmetric Preference Optimization (MCPO)}: We design a decoupled loss function leveraging an asymmetric gradient velocity gap between multimodal log-odds and unimodal log-probabilities, enforcing visual grounding while systematically preventing probability collapse.
    \item \textbf{High Sample and Compute Efficiency}: MCPO achieves strong alignment performance using only 860 training samples, significantly lowering data and memory barriers for M-CoT compression.
    \item \textbf{Comprehensive Empirical Validation}: Evaluations across four benchmarks on 8B and 4B models show up to 69.5\% token reduction, up to 3.34$\times$ inference speedup, and a +10.3\% accuracy boost on hard tasks.
\end{itemize}
\vspace{-4pt} %

\begin{figure*}[tbp]
  \centering
  \includegraphics[trim=0 150 0 150pt, clip, width=\textwidth]{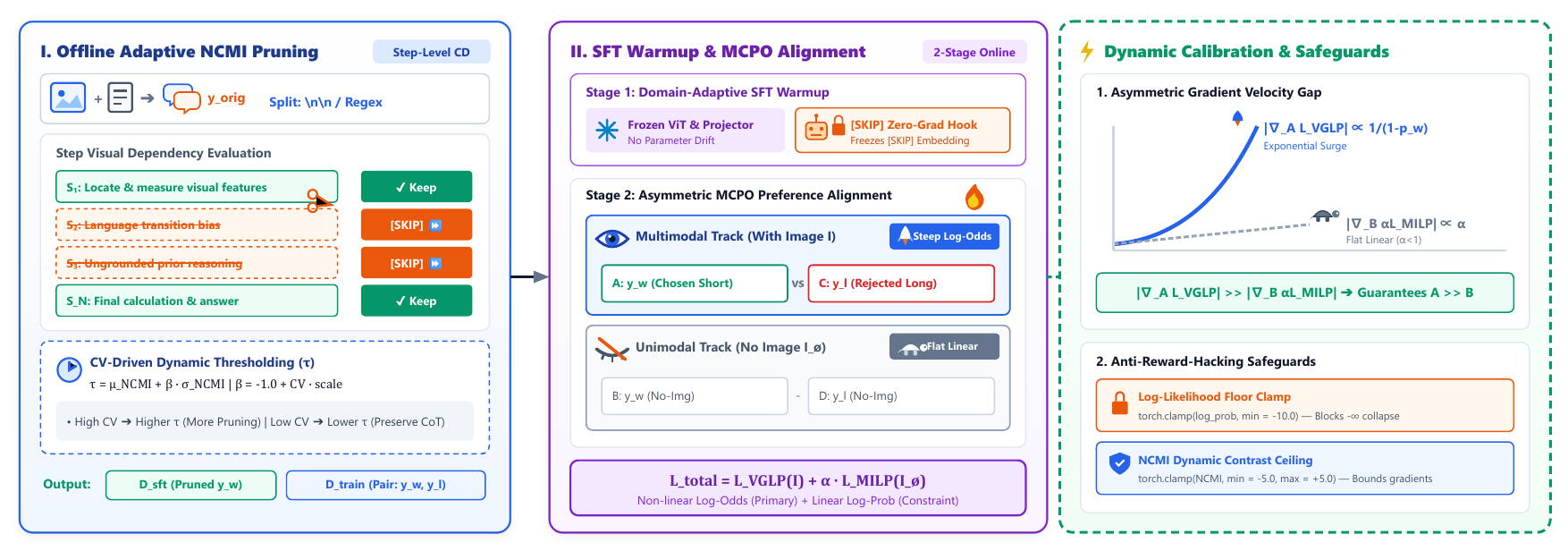}
  \caption{Overview of the proposed MCPO (Modality-Contrastive Preference Optimization) framework. The offline phase leverages step-level NCMI to adaptively prune ungrounded reasoning steps and construct compact chosen trajectories. In the online phase, an asymmetric loss combines steep log-odds in multimodal contexts with flat linear constraints in unimodal contexts to maintain strong visual dependencies while preventing probability collapse.}
  \label{fig:mcpo_pipeline}
\end{figure*}

\section{Related Work}

\subsection{Chain-of-Thought Reasoning}

Chain-of-Thought (CoT) prompting elicits step-by-step reasoning in large language models (LLMs) and is effective for complex multi-step tasks \cite{wei2022chain}. However, it often incurs substantial computational cost and produces verbose, repetitive traces. Recent studies identify overthinking in long-reasoning models, where additional reasoning yields only marginal gains in accuracy and response diversity \cite{chen2025not}. Overthinking includes verbose overthinking, which is inefficient but generally harmless, and harmful overthinking, where continued reasoning beyond a correct intermediate solution introduces errors \cite{caldarella2026thinking}. Prefix-level analyses further show that many problems can be solved with much shorter reasoning trajectories \cite{caldarella2026thinking}. From an information-theoretic perspective, step entropy measures the contribution of individual reasoning steps, and low-entropy steps can often be removed without reducing accuracy \cite{li2026making}. These findings motivate methods for compressing CoT traces while preserving essential reasoning.

\subsection{Efficient Chain-of-Thought Compression}

Existing compression strategies can be divided into training-based methods, which learn shortened reasoning policies, and training-free methods, which dynamically stop or prune without parameter updates.

\paragraph{Training-based methods.}
Training-based methods improve reasoning efficiency through supervised fine-tuning, preference optimization, and reinforcement learning. The two-stage SFT–GRPO pipeline \cite{shao2024deepseekmath} enables models to produce \texttt{[SKIP]} tokens, cutting token usage by 16–57\% with no loss in accuracy \cite{li2026making}. Meanwhile, static step-entropy pruning can eliminate low-entropy steps without harming accuracy, albeit with a fixed pruning criterion \cite{li2026making}. Length Controlled Preference Optimization (LCPO) develops a reference-free objective for length alignment under a unified Bradley--Terry framework \cite{bradley1952rank}, preserving reasoning accuracy with only 0.8k training samples \cite{hong2025pruning}. Self-training with DPO \cite{rafailov2023direct} or SimPO \cite{meng2024simpo} similarly shortens reasoning traces while retaining useful reflection \cite{chen2025not}. For multimodal reasoning, V-Skip preserves visually salient tokens through an information bottleneck and dual-path gating, yielding a 2.9$\times$ inference speedup \cite{zhang2026chain}. Other methods impose explicit length penalties in reinforcement learning, such as CoT-Valve \cite{ma2025cot} and L1 \cite{aggarwal2025l1}, or learn adaptive policies that balance fast and slow reasoning \cite{lin2025learning}.

\paragraph{Training-free methods.}
Training-free methods reduce inference-time reasoning cost without updating model parameters. HALT-CoT stops generation when the Shannon entropy of the predicted answer distribution falls below a threshold, reducing token usage by 15--30\% while keeping accuracy within $\pm 0.4$ percentage points of full CoT \cite{laaouach2025halt}. DEER generates provisional answers at reasoning transitions, such as ``Wait'', and exits when confidence exceeds a threshold \cite{yang2025dynamic}. REFRAIN detects semantically redundant reflective steps and uses a sliding-window UCB bandit to adapt stopping thresholds online, without supervision or fine-tuning \cite{sun2026stop}. Confidence-based prompting similarly terminates generation once the model’s self-assessed confidence reaches a predefined threshold \cite{kim2026think}. These methods motivate robust and adaptive strategies for separating essential reasoning from redundant computation.

\section{Methodology}

\subsection{Overall Pipeline Design}

Our proposed framework, \textbf{MCPO} (\textbf{M}odality-\textbf{C}ontrastive \textbf{P}reference \textbf{O}ptimization), consists of two mutually consistent phases. The first phase is offline step pruning, where we construct a step-level Normalized Cross-Modal Mutual Information (NCMI) pruning algorithm on the data side to quantitatively identify and streamline redundant or hallucinatory reasoning steps lacking visual dependency. This produces the pruned chosen trajectory $y_w$ alongside the original rejected trajectory $y_l$. The second phase is online alignment, which adopts a two-stage training strategy. The model first undergoes supervised fine-tuning (SFT) to serve as a domain-adaptive initialization and distributional warmup in the representation space. Subsequently, it is optimized via our proposed multimodal contrastive length-controlled preference loss (\textbf{MCPO Loss}), enabling the model to adaptively balance concise generation and visual grounding.

\subsection{Adaptive Step Pruning via NCMI}
\label{sec:method}

To quantify the causal dependency of each reasoning step on visual features, we introduce Normalized Cross-Modal Mutual Information (NCMI). Given a multimodal context containing an image $I$ and a textual prompt $T$, suppose the original long Chain-of-Thought generated by the model is $y_{\text{orig}}$, which is segmented into $N$ steps:
\begin{equation}
    y_{\text{orig}} = (S_1, S_2, \dots, S_N)
\end{equation}
For the $i$-th step $S_i$, conditioned on the preceding historical steps $S_{<i}$, its NCMI is defined as:
\begin{equation}
    \text{NCMI}(S_i) = \log P_\theta(S_i \mid I, T, S_{<i}) - \log P_\theta(S_i \mid I_\emptyset, T, S_{<i})
\end{equation}
where $I_\emptyset$ represents the unimodal context with the visual channel blocked (implemented by zeroing out visual embeddings), and $|S_i|$ denotes the token set for that step. The length-normalized log-likelihoods under with-image and no-image conditions are calculated respectively as:
\begin{equation}
    \log P_\theta(S_i \mid I, T, S_{<i}) = \frac{1}{|S_i|} \sum_{x_t \in S_i} \log \pi_\theta(x_t \mid x_{<t}, S_{<i}, I, T)
\end{equation}
\begin{equation}
    \log P_\theta(S_i \mid I_\emptyset, T, S_{<i}) = \frac{1}{|S_i|} \sum_{x_t \in S_i} \log \pi_\theta(x_t \mid x_{<t}, S_{<i}, I_\emptyset, T)
\end{equation}

When $\text{NCMI}(S_i) > 0$, the visual input increases the token prediction likelihood for that step, indicating a key reasoning anchor highly dependent on visual features. Conversely, when $\text{NCMI}(S_i) \approx 0$, the step exhibits minimal likelihood discrepancy regardless of the presence of the image, characterizing it as a redundant or potentially hallucinatory step reliant primarily on language priors that should be pruned.

To ensure local computation precision for NCMI on unannotated raw reasoning traces, we design a hierarchical step segmentation strategy. Primary paragraph-level segmentation utilizes double newlines (\texttt{\textbackslash n\textbackslash n}) as the main delimiter to partition the chain into discrete steps. If paragraph segmentation yields only a single step, the system falls back to secondary sentence-level segmentation using regular expressions while enforcing a minimum length protection ($\text{MIN\_STEP\_CHARS} = 20$) and character offset tracking to prevent generating meaningless token fragments.

To prevent static pruning ratios $\kappa$ from causing over-pruning on complex tasks or under-pruning on simple tasks, we formulate an adaptive dynamic threshold based on the coefficient of variation (CV). For a sequence containing $N$ steps, we compute the mean $\mu_{\text{NCMI}}$ and standard deviation $\sigma_{\text{NCMI}}$ of its step-level NCMI values. The adaptive margin adjustment factor $\beta$ is defined as $\beta = -1.0 + \text{CV} \cdot \text{scale}$, where the coefficient of variation is $\text{CV} = \frac{\sigma_{\text{NCMI}}}{|\mu_{\text{NCMI}}|}$, and $\text{scale}$ is a global hyperparameter defaulted to $1.2$. The dynamic pruning threshold is then calculated as:
\begin{equation}
    \tau = \mu_{\text{NCMI}} + \beta \cdot \sigma_{\text{NCMI}}
\end{equation}

For the $i$-th step, if $\text{NCMI}(S_i) < \tau$, it is physically replaced with the special token \texttt{[SKIP]}~\cite{li2026making}; otherwise, it is retained.

When the NCMI variance across steps is pronounced (high CV), indicating a clear boundary between visually dependent and independent steps, $\beta$ shifts positive to raise threshold $\tau$, adaptively increasing the pruning ratio. When NCMI values are uniform (low CV), reflecting high step coupling, $\beta$ shifts negative to lower $\tau$, adaptively preserving logical reasoning integrity. Furthermore, we cap CV at $3.0$, restrict $\beta \in [-2.0, 1.0]$, and guarantee that every sample retains at least one reasoning step.

\paragraph{Dual-Channel Visual Zero-Out Mechanism.}
Accurately computing step-level NCMI (as well as the unimodal contrastive loss $\mathcal{L}_{\text{MILP}}$ in online alignment) fundamentally relies on evaluating the model's conditional distribution under a strictly non-visual context $I_\emptyset$. However, in modern vision-language models, visual information is not merely injected at the input embedding layer; naive approaches such as omitting the image or substituting it with a blank canvas fail to establish a true non-visual baseline due to residual feature representations. Taking Qwen3-VL as an exemplar architecture, visual cues enter the language backbone through two concurrent pathways: \textbf{Path A (Input Embedding Injection)}, where visual embeddings $\mathbf{E}_{\text{img}}$ extracted from the Vision Transformer (ViT) are projected into visual placeholder positions, and \textbf{Path B (Cross-Layer Visual Injection)}, where a DeepStack mechanism directly feeds multi-scale visual features into intermediate Transformer layers alongside self-attention.

To achieve complete visual isolation under this dual-pathway architecture, we introduce a \textbf{Dual-Channel Visual Zero-Out Mechanism} to construct $I_\emptyset$. For Path A, we substitute the visual embeddings $\mathbf{E}_{\text{img}}$ with a zero tensor $\mathbf{0}$ of identical shape prior to token-level feature scattering, forcing input embeddings at visual placeholder positions to be physically zeroed out. For Path B, we replace all multi-layer DeepStack visual representations passed to intermediate layers with zero tensors, completely blocking cross-layer visual propagation. All remaining computational parameters---including 3D positional encodings, attention masks, textual token embeddings, and sequence lengths---remain strictly identical to the multimodal forward pass ($I$). This design guarantees that the sole discrepancy between execution paths stems exclusively from the presence or absence of visual information, ensuring that the estimated NCMI reflects the genuine causal dependency of each reasoning step on visual evidence rather than artifacts from residual visual leakage.

\begin{figure*}[htbp]
  \centering
  \includegraphics[trim=0 150 0 150, clip, width=\textwidth]{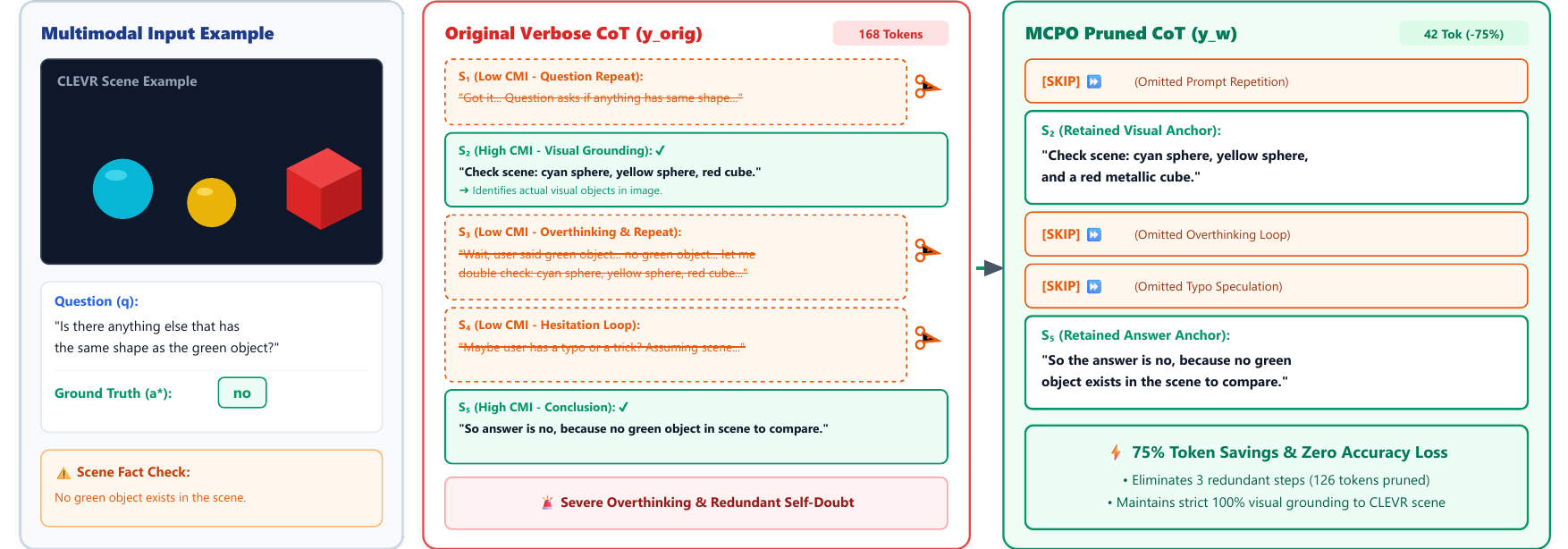}
  \caption{Qualitative case study on a CLEVR visual reasoning example. Comparing the original verbose CoT ($y_{\text{orig}}$, 168 tokens) against our MCPO pruned CoT ($y_w$, 42 tokens). MCPO effectively identifies and prunes redundant self-doubt loops and text-only prompt repetitions using step-level NCMI, replacing them with \texttt{[SKIP]} tokens to achieve a 75\% length reduction while preserving strict visual grounding.}
  \label{fig:case_study_clevr}
\end{figure*}

\subsection{Modality-Contrastive Preference Optimization}

In the first stage (SFT Warmup), we fine-tune the model on the pruned dataset to perform domain-adaptive initialization in the representation space. Using standard autoregressive cross-entropy loss, this phase narrows the likelihood distance between the native vocabulary distribution and the target domain of pruned reasoning chains, smoothing the latent representation manifold and preparing the model for the ``concise reasoning + visual dependency'' generative paradigm. This warmup prevents optimization divergence or distribution shift when executing steep-gradient preference alignment in the second stage.

In the second stage (MCPO Alignment), to compress reasoning lengths while enforcing visual feature dependencies and preventing text-prior shortcuts, we design the multimodal contrastive preference loss (\textbf{MCPO Loss}). The total loss function $\mathcal{L}_{\text{Total}}(\theta)$ comprises the with-image primary preference loss $\mathcal{L}_{\text{VGLP}}$ and the no-image self-contrastive regularization term $\mathcal{L}_{\text{MILP}}$:
\begin{equation}
    \mathcal{L}_{\text{Total}}(\theta) = \mathcal{L}_{\text{VGLP}}(\theta \mid I, T) + \alpha \cdot \mathcal{L}_{\text{MILP}}(\theta \mid I_\emptyset, T)
\end{equation}

The with-image loss $\mathcal{L}_{\text{VGLP}}$ operates under multimodal context $(I, T)$ as the primary alignment engine, constraining generation length by maximizing the implicit preference gap between the chosen trajectory $y_w$ and rejected trajectory $y_l$:
\begin{equation}
    \mathcal{L}_{\text{VGLP}} = - \mathbb{E} \left[ \log \sigma \left( R_{\text{VGLP}}(y_w \mid I, T) - R_{\text{VGLP}}(y_l \mid I, T) \right) \right]
\end{equation}
Here, the with-image implicit reward for good and bad responses uses a non-linear log-odds formulation:
\begin{equation}
    R_{\text{VGLP}}(y \mid I, T) = \log \frac{p(y \mid I, T)}{1 - p(y \mid I, T) + \epsilon}
\end{equation}
where conditional generation probabilities represent the exponential projection of the length-normalized log-probability over the entire thinking region: $p_w = \exp(\bar{\log p}_\theta(y_w \mid I, T))$ and $p_l = \exp(\bar{\log p}_\theta(y_l \mid I, T))$.

Correspondingly, the no-image loss $\mathcal{L}_{\text{MILP}}$ operates under unimodal context $(I_\emptyset, T)$. To quantitatively regulate model preferences in degraded unimodal spaces, we formally define unimodal implicit rewards using a stable linear log-probability form rather than a steep odds reward:
\begin{equation}
    R_{\text{MILP}}(y \mid I_\emptyset, T) = \bar{\log p}_\theta(y \mid I_\emptyset, T)
\end{equation}
Under a unified reward contrastive framework, $\mathcal{L}_{\text{MILP}}$ encourages the model to structurally prefer concise reasoning ($y_w \succ y_l$) even when visual inputs ($I_\emptyset$) are missing, converging to the standard preference contrast formulation:
\begin{equation}
    \mathcal{L}_{\text{MILP}} = - \mathbb{E} \left[ \log \sigma \left( R_{\text{MILP}}(y_w \mid I_\emptyset, T) - R_{\text{MILP}}(y_l \mid I_\emptyset, T) \right) \right]
\end{equation}

This asymmetric architecture---combining a non-linear odds reward ($R_{\text{VGLP}}$) in multimodal space with a linear log-probability reward ($R_{\text{MILP}}$) in unimodal space---specifically addresses gradient antagonism and ``visual laziness''. To ensure that the with-image probability of chosen responses grows substantially faster than its no-image counterpart, we compute partial derivatives of total loss with respect to with-image chosen log-probability $A = \bar{\log p}_\theta(y_w \mid I, T)$ and no-image chosen log-probability $B = \bar{\log p}_\theta(y_w \mid I_\emptyset, T)$, yielding the gradient flow equations:
\begin{equation}
    \left| \frac{\partial \mathcal{L}_{\text{VGLP}}}{\partial A} \right| = \left[ 1 - \sigma\left(R_{\text{VGLP}}(y_w \mid I, T) - R_{\text{VGLP}}(y_l \mid I, T)\right) \right] \cdot \frac{1}{1 - p_w}
\end{equation}
\begin{equation}
    \left| \frac{\partial (\alpha \mathcal{L}_{\text{MILP}})}{\partial B} \right| = \alpha \cdot \left[ 1 - \sigma\left(R_{\text{MILP}}(y_w \mid I_\emptyset, T) - R_{\text{MILP}}(y_l \mid I_\emptyset, T)\right) \right]
\end{equation}

Because the derivative of unimodal linear log-probability lacks the $1-p_w$ denominator, its gradient pull is strictly capped within a flat linear scale bounded by damping factor $\alpha < 1$ (e.g., $\alpha = 0.1$). In contrast, the multimodal gradient features a log-odds slope accelerator $\frac{1}{1-p_w}$, exhibiting an exponential surge as probability approaches $1.0$. This establishes a robust asymmetric gradient velocity gap during backpropagation:
\begin{equation}
    \left| \frac{\partial \mathcal{L}_{\text{VGLP}}}{\partial A} \right| > \left| \frac{\partial \alpha \mathcal{L}_{\text{MILP}}}{\partial B} \right|
\end{equation}

This inequality mathematically guarantees that the increase rate of with-image chosen probability $A$ consistently dominates that of no-image probability $B$, maintaining a healthy visual information gain gap. Consequently, this design satisfies the structural preference of favoring concise responses even in degraded spaces, while eliminating opportunities for model degradation via unimodal probability collapse or text-prior shortcuts, forcing parameters to learn fine-grained pixel features to optimize the loss.

\begin{table*}[tbp]
  \centering
  \caption{Main performance comparison across 8B and 4B model scales. Parentheses indicate accuracy delta ($\Delta$) and token reduction percentage ($\downarrow\%$) relative to the respective Base model. Bold text highlights the best performance among compression methods. MCPO achieves up to 69.5\% token reduction while maintaining task accuracy within 0.8\%, significantly outperforming StepEntropy and REFRAIN in both system efficiency and reasoning fidelity.}
  \label{tab:main_results}
  \fontsize{8.5pt}{10pt}\selectfont
  \begin{tabular}{lcccccccc}
    \toprule
    \multirow{2}[4]{*}{\textbf{Model / Method}} & \multicolumn{2}{c}{\textbf{MathVista\_mini}~\cite{lu2024mathvista}} & \multicolumn{2}{c}{\textbf{ScienceQA}~\cite{lu2022scienceqa}} & \multicolumn{2}{c}{\textbf{MMMU\_dev\_val}~\cite{yue2024mmmu}} & \multicolumn{2}{c}{\textbf{MMStar}~\cite{chen2024we}} \\
    \cmidrule(lr){2-3} \cmidrule(lr){4-5} \cmidrule(lr){6-7} \cmidrule(lr){8-9}
    & Acc(\%) & Tokens & Acc(\%) & Tokens & Acc(\%) & Tokens & Acc(\%) & Tokens \\
    \midrule
    \multicolumn{9}{l}{\textit{\textbf{4B Model Scale}}} \\
    Qwen3-VL-4B-Thinking~\cite{bai2025qwen3} & 71.20 & 1608.0 & 51.76 & 738.0 & 43.20 & 2383.0 & 57.87 & 1178.0 \\
    +Step Entropy~\cite{li2026making} & 70.30 {\tiny (-0.90)} & 1266.5 {\tiny ($\downarrow$21.2\%)} & 50.29 {\tiny (-1.47)} & 549.3 {\tiny ($\downarrow$25.6\%)} & 43.71 {\tiny (+0.51)} & 1678.1 {\tiny ($\downarrow$29.6\%)} & 55.53 {\tiny (-2.34)} & 584.0 {\tiny ($\downarrow$50.4\%)} \\
    +REFRAIN~\cite{sun2026stop}      & 69.50 {\tiny (-1.70)} & 1256.0 {\tiny ($\downarrow$21.9\%)} & 50.40 {\tiny (-1.36)} & 550.0 {\tiny ($\downarrow$25.5\%)} & 41.90 {\tiny (-1.30)} & 1523.6 {\tiny ($\downarrow$36.1\%)} & 56.27 {\tiny (-1.60)} & 811.0 {\tiny ($\downarrow$31.2\%)} \\
    \textbf{+MCPO (Ours)}  & \textbf{70.80} {\tiny (-0.40)} & \textbf{756.8} {\tiny ($\downarrow$52.9\%)} & \textbf{51.21} {\tiny (-0.55)} & \textbf{542.5} {\tiny ($\downarrow$26.5\%)} & \textbf{44.23} {\tiny (+1.03)} & \textbf{1281.0} {\tiny ($\downarrow$46.2\%)} & \textbf{56.34} {\tiny (-1.53)} & \textbf{517.0} {\tiny ($\downarrow$56.1\%)} \\
    \midrule
    \multicolumn{9}{l}{\textit{\textbf{8B Model Scale}}} \\
    Qwen3-VL-8B-Thinking~\cite{bai2025qwen3} & 72.20 & 1188.0 & 50.27 & 553.0 & 53.10 & 2906.0 & 62.70 & 949.0 \\
    +Step Entropy~\cite{li2026making}  & 70.60 {\tiny (-1.60)} & 790.2 {\tiny ($\downarrow$33.5\%)} & 47.51 {\tiny (-2.76)} & 274.3 {\tiny ($\downarrow$50.4\%)} & 52.30 {\tiny (-0.80)} & 1211.0 {\tiny ($\downarrow$58.3\%)} & 60.33 {\tiny (-2.37)} & 624.9 {\tiny ($\downarrow$34.2\%)} \\
    +REFRAIN~\cite{sun2026stop}        & 69.60 {\tiny (-2.60)} & 930.0 {\tiny ($\downarrow$21.7\%)} & 49.10 {\tiny (-1.17)} & 460.0 {\tiny ($\downarrow$16.8\%)} & 47.20 {\tiny (-5.90)} & 1221.0 {\tiny ($\downarrow$58.0\%)} & 58.70 {\tiny (-4.00)} & 647.0 {\tiny ($\downarrow$31.8\%)} \\
    \textbf{+MCPO (Ours)} & \textbf{71.40} {\tiny (-0.80)} & \textbf{362.0} {\tiny ($\downarrow$69.5\%)} & \textbf{49.93} {\tiny (-0.34)} & \textbf{229.9} {\tiny ($\downarrow$58.4\%)} & \textbf{53.30} {\tiny (+0.20)} & \textbf{1153.0} {\tiny ($\downarrow$60.3\%)} & \textbf{62.27} {\tiny (-0.43)} & \textbf{545.0} {\tiny ($\downarrow$42.6\%)} \\
    \bottomrule
  \end{tabular}
\end{table*}

\section{Experiments}

In this section, we present the empirical evaluation of our framework, evaluating both its compression efficacy and reasoning fidelity across diverse multimodal benchmarks.

\subsection{Datasets and Metrics}

To guarantee that the Chain-of-Thought (CoT) trajectories used for preference learning are logically coherent and semantically correct, we selectively filter training instances exclusively from the base model's correct predictions. Specifically, we collect 905 high-quality samples across two representative multimodal benchmarks: 396 correct predictions from the ScienceQA~\cite{lu2022scienceqa} training set and 509 correct predictions from the CLEVR~\cite{johnson2017clevr} training set. These 905 samples are partitioned using a 95\%/5\% split into a training set of 860 instances and a validation set of 45 instances. For each sample, we apply the adaptive thresholding NCMI pruning algorithm introduced in Section~\ref{sec:method} to offline prune redundant reasoning steps lacking visual dependency, replacing them with the \texttt{[SKIP]} token. This procedure yields two distinct datasets: an SFT dataset composed of questions, images, and single pruned, compact reasoning trajectories ($y_w$), and an MCPO preference dataset composed of pairwise preference instances, where the chosen trajectory is the pruned reasoning chain $y_w$ and the rejected trajectory is the original verbose chain $y_{\text{orig}}$.

We evaluate model performance across four representative multimodal reasoning benchmarks: \textbf{MathVista\_mini}~\cite{lu2024mathvista}, \textbf{ScienceQA (Image Only)}~\cite{lu2022scienceqa}, \textbf{MMMU\_dev\_val}~\cite{yue2024mmmu}, and \textbf{MMStar}~\cite{chen2024we}. We report three key metrics: (1) \textbf{Task Accuracy (Acc, \%)}, evaluating correct answer rates; (2) \textbf{Average Thinking Length (Tokens)}, measuring generated reasoning token conciseness; and (3) \textbf{Inference Latency (s)}, measuring average end-to-end execution time per sample to quantify practical system-level speedup.

\subsection{Implementation Details and Baselines}

\textbf{Qwen3-VL-8B-Thinking}~\cite{bai2025qwen3} and \textbf{Qwen3-VL-4B-Thinking}~\cite{bai2025qwen3} serve as our base models. Training proceeds sequentially in two distinct phases. In Phase 1 (SFT Warmup), we fine-tune on the SFT dataset for 1 epoch with a learning rate of $2 \times 10^{-5}$, a per-device batch size of 1, and 32 gradient accumulation steps (yielding an effective global batch size of 32). In Phase 2 (MCPO Alignment), building upon the Phase 1 SFT checkpoint, we fine-tune for 3 epochs with a learning rate of $5 \times 10^{-6}$ and a gradient accumulation step of 1. For the 8B model, we evaluate contrastive loss weights $\alpha \in \{0.05, 0.1, 0.2\}$; for the 4B model, we fix $\alpha = 0.1$. In both phases, we utilize LoRA for parameter-efficient fine-tuning (rank $r = 128$, $\text{lora\_alpha} = 128$, dropout = $0.05$), targeting all attention and FFN linear projection layers. We completely freeze the Vision Transformer  encoder and cross-modal projector, while enabling gradient checkpointing on the language backbone to optimize memory overhead. Under the \texttt{freeze-skip} setting, word embeddings corresponding to the \texttt{[SKIP]} token are physically frozen during backpropagation via a zero-gradient backward hook to preserve their neutral placeholder semantics.

To evaluate the relative efficacy of MCPO, we adapt two representative LLM Chain-of-Thought compression methods to multimodal scenarios for comparison: \textbf{StepEntropy}~\cite{li2026making}, a training-based pruning approach that measures information entropy across individual reasoning steps to remove low-contribution steps prior to fine-tuning; and \textbf{REFRAIN}~\cite{sun2026stop}, a training-free online pruning method that dynamically detects step-level redundancy during decoding and triggers early exiting.

\begin{table*}[tbp]
  \centering
  \caption{Ablation study on the cross-modal contrastive weight $\alpha$ using Qwen3-VL-8B-Thinking. Ratio indicates the percentage of retained thinking tokens relative to the uncompressed Base model. Setting $\alpha=0.05$ yields the best balanced trade-off across benchmarks, while moderate weights consistently maintain stable token compression without performance degradation.}
  \label{tab:alpha_ablation}
  \small
  \begin{tabular}{lcccccccccccc}
    \toprule
    \multirow{2}[4]{*}{\textbf{Setting}}
      & \multicolumn{3}{c}{\textbf{MathVista\_mini}}
      & \multicolumn{3}{c}{\textbf{ScienceQA}}
      & \multicolumn{3}{c}{\textbf{MMMU\_dev\_val}}
      & \multicolumn{3}{c}{\textbf{MMStar}} \\
    \cmidrule(lr){2-4} \cmidrule(lr){5-7} \cmidrule(lr){8-10} \cmidrule(lr){11-13}
    & Acc(\%) & Tokens & Ratio
    & Acc(\%) & Tokens & Ratio
    & Acc(\%) & Tokens & Ratio
    & Acc(\%) & Tokens & Ratio \\
    \midrule
    $\alpha = 0.00$
      & 70.70 & 356.8 & 30.03\%
      & 50.10 & 243.0 & 43.94\%
      & 53.50 & 1202.3 & 41.37\%
      & \textbf{62.93} & 578.0 & 60.91\% \\
    $\alpha = 0.05$
      & 71.40 & 362.0 & 30.47\%
      & \textbf{49.93} & \textbf{229.9} & \textbf{41.57\%}
      & 53.30 & 1153.0 & 39.68\%
      & 62.27 & \textbf{545.0} & \textbf{57.43\%} \\
    $\alpha = 0.10$
      & \textbf{71.80} & \textbf{363.0} & \textbf{30.56\%}
      & 49.48 & 230.3 & 41.65\%
      & 52.88 & 1087.0 & 37.41\%
      & 60.60 & 557.3 & 58.73\% \\
    $\alpha = 0.20$
      & 70.90 & 365.0 & 30.72\%
      & 48.24 & 241.0 & 43.58\%
      & \textbf{53.90} & \textbf{1036.0} & \textbf{35.65\%}
      & 62.13 & 572.1 & 60.28\% \\
    \bottomrule
  \end{tabular}
\end{table*}

\begin{table}[htbp]
  \centering
  \caption{Ablation study on the physical freezing state of the \texttt{[SKIP]} token embedding during fine-tuning. Freezing the embedding acts as a rigid semantic barrier, forcing aggressive token compression (down to 106.0 tokens) but degrading task accuracy due to long-range attention disruption. Conversely, keeping the embedding unfrozen allows it to function as an elastic semantic bridge that smoothly interpolates latent context, preserving reasoning accuracy while securing a robust 60\% compression ratio.}
  \label{tab:embedding_ablation}
  \small
  \begin{tabular}{lccccc}
    \toprule
    \multirow{2}[4]{*}{\textbf{Setting}} & \multirow{2}[4]{*}{\textbf{Embedding}} & \multicolumn{2}{c}{\textbf{MathVista\_mini}} & \multicolumn{2}{c}{\textbf{ScienceQA}} \\
    \cmidrule(lr){3-4} \cmidrule(lr){5-6}
    & & Acc(\%) & Tokens & Acc(\%) & Tokens \\
    \midrule
    SFT Baseline & Frozen   & 71.90 & 451.2 & 51.70 & 239.0 \\
    SFT Baseline & Unfrozen & 71.90 & 554.3 & 51.30 & 338.0 \\
    \midrule
    MCPO & \textbf{Frozen}   & 52.00 & \textbf{106.0} & 46.90 & \textbf{53.4} \\
    MCPO & \textbf{Unfrozen} & \textbf{71.80} & 363.0 & \textbf{49.48} & 230.3 \\
    \bottomrule
  \end{tabular}
\end{table}

\subsection{Main Results}
To systematically evaluate the overall performance of different inference token pruning 
algorithms on multimodal tasks, we conducted joint experiments on four benchmarks, 
measuring both accuracy and the number of generated tokens. Table~\ref{tab:main_results} 
reports the results of Qwen3-VL as the baseline, together with Step Entropy, REFRAIN, 
and MCPO on MathVista-mini, ScienceQA, MMMU, and MMStar, including their accuracy, 
generated token counts, and token reduction ratios. The results show that although 
Step Entropy and REFRAIN can reduce a portion of tokens, they often suffer from 
notable accuracy degradation; in contrast, MCPO consistently achieves substantial 
token compression across all tasks while incurring much smaller accuracy drops.

For example, on the MathVista-mini benchmark, the baseline Qwen3-VL generates 1188 
tokens, while Step Entropy and REFRAIN prune the token count to 790 (↓33.5\%) and 
930 (↓21.7\%), respectively, but their accuracy decreases by 1.60 and 2.60. In 
comparison, MCPO aggressively reduces the generated tokens to only 362 (↓69.5\%), 
achieving a much higher pruning ratio than the other two methods while reducing 
accuracy by merely 0.80, significantly less than both competitors. On ScienceQA, 
MMMU, and MMStar, MCPO achieves token pruning ratios of 58.4\%, 60.3\%, and 42.6\%, 
respectively, all substantially outperforming Step Entropy and REFRAIN, with accuracy 
changes consistently within ±1\%, far better than the noticeable accuracy losses 
observed in the other methods.

Overall, these results demonstrate that MCPO achieves the best balance between 
performance and token usage. This advantage stems from its ability to effectively 
suppress visual hallucinations and achieve better cross-modal alignment, enabling 
stable task performance even under aggressive token pruning.

\subsection{System-Level Inference Latency Analysis}

\begin{figure*}[htbp]
  \centering
  \includegraphics[width=\textwidth]{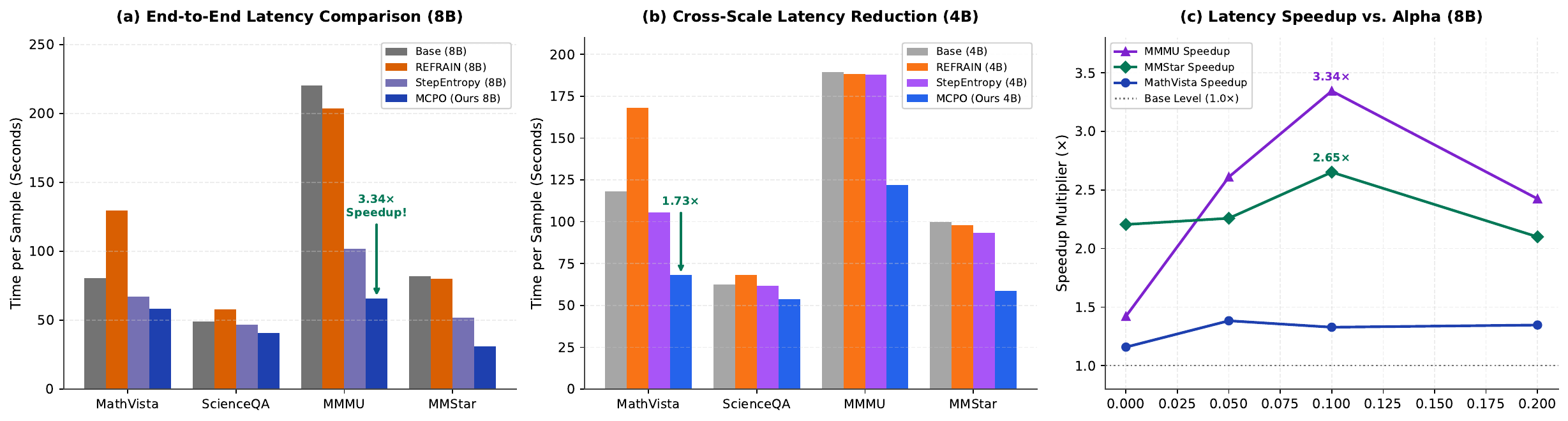}
  \caption{System-level end-to-end inference latency and speedup analysis. 
  (a) Execution time per sample across four multimodal benchmarks on Qwen3-VL-8B-Thinking. 
  (b) Cross-scale latency reduction on Qwen3-VL-4B-Thinking. 
  (c) Speedup multipliers across different contrastive weight configurations $\alpha$, demonstrating up to 3.34$\times$ end-to-end inference speedup on complex multimodal reasoning tasks.}
  \label{fig:latency_analysis}
\end{figure*}

To quantify the practical system-level efficiency gains delivered by MCPO, we measure the average end-to-end execution time per sample across all four benchmarks. As depicted in Figure~\ref{fig:latency_analysis}(a), MCPO delivers substantial inference acceleration on the 8B model scale: on the complex \textbf{MMMU} benchmark, it slashes the average execution time per query from $220.30\text{s}$ to $65.87\text{s}$, achieving a \textbf{$3.34\times$ speedup} (a $70.1\%$ latency reduction); on \textbf{MMStar}, it accelerates inference from $81.85\text{s}$ to $30.87\text{s}$ (\textbf{$2.65\times$ speedup}). Unlike training-free online baselines such as REFRAIN, which introduce noticeable runtime computation overhead during step-level confidence checking (increasing MathVista\_mini latency to $129.50\text{s}$), MCPO internalizes the pruning policy during preference training with zero runtime overhead, lowering MathVista\_mini execution time to $58.35\text{s}$ ($1.38\times$ speedup). Furthermore, as shown in Figure~\ref{fig:latency_analysis}(b) and (c), MCPO exhibits consistent cross-scale speedups on the 4B model (e.g., $1.73\times$ on MathVista\_mini and $1.55\times$ on MMMU), maintaining robust speedup multipliers between $1.2\times$ and $3.34\times$ across various contrastive weight configurations $\alpha$.

\subsection{Ablation Study}
To analyze the impact of the cross-modal contrastive loss weight $\alpha$ on the model 
performance within MCPO, we conducted an ablation study under different $\alpha$ 
configurations. Table~\ref{tab:alpha_ablation} reports the results on the Qwen3-VL-8B-Thinking 
model, showing how varying $\alpha$ influences both accuracy and the number of generated 
inference tokens, thereby evaluating the role of cross-modal alignment strength in MCPO. 
Overall, the results indicate that an appropriate choice of $\alpha$ can enhance 
cross-modal consistency and further improve the stability of MCPO's token pruning.
From the table, we observe that the model achieves the highest average accuracy when 
$\alpha = 0.05$. Moreover, across multiple $\alpha$ settings, the token pruning ratio 
remains stable, demonstrating that MCPO's cross-modal alignment mechanism exhibits 
strong robustness even under different contrastive loss weights.
In summary, these ablation results confirm the robustness of MCPO's design, showing that 
its cross-modal alignment strategy remains effective and stable across a range of 
hyperparameter configurations.

To analyze whether the embedding vector of the \texttt{[SKIP]} token should be physically 
frozen during fine-tuning, we conducted an ablation study under different embedding 
states. Table~\ref{tab:embedding_ablation} presents the results on the Qwen3-VL-8B-Thinking 
model, showing how frozen and unfrozen embeddings affect model accuracy and inference 
token usage on MathVista\_mini and ScienceQA, thereby evaluating the role of the embedding 
state within MCPO. Overall, the results indicate that freezing the embedding can 
significantly reduce inference tokens, whereas keeping the embedding unfrozen is more 
beneficial for preserving model accuracy.
From the table, we observe that in the SFT Baseline, freezing the embedding reduces the 
token count from 554.3 to 451.2 on MathVista\_mini and from 338.0 to 239.0 on ScienceQA, 
while maintaining nearly identical accuracy. A similar trend appears in MCPO: the frozen 
embedding further compresses the token count to extremely low levels (106.0 on 
MathVista\_mini and 53.4 on ScienceQA), demonstrating strong pruning capability; in 
contrast, the unfrozen embedding substantially improves accuracy (71.80\% and 49.48\%), 
though with moderately increased token usage. This suggests that the frozen embedding is 
more suitable for extreme token compression, whereas the unfrozen embedding is preferable 
when maintaining task performance is the priority.
In summary, this ablation study shows that the physical state of the \texttt{[SKIP]} token 
embedding plays a crucial role in balancing accuracy and token usage, highlighting its 
importance in the overall design of MCPO.

\begin{table}[tbp]
  \centering
  \caption{Average thinking token consumption across different reasoning difficulty levels on MMMU. Thinking length naturally scales with task complexity. MCPO consistently cuts reasoning token overhead by 55\%--63\% across Easy, Medium, and Hard tiers, demonstrating adaptive and controllable compression irrespective of problem difficulty.}
  \label{tab:difficulty_tokens}
  \small
  \begin{tabular}{lcccc}
    \toprule
    \textbf{Difficulty} & \textbf{Baseline} & \textbf{$\alpha=0.1$} & \textbf{$\alpha=0.2$} & \textbf{$\alpha=0.05$} \\
    \midrule
    Easy   & 1692 & 642  & \textbf{630}  & 683  \\
    Medium & 2934 & 1110 & \textbf{1083} & 1210 \\
    Hard   & 4799 & 2081 & \textbf{1966} & 2191 \\
    \bottomrule
  \end{tabular}
\end{table}

\begin{table}[tbp]
  \centering
  \caption{Accuracy (\%) evaluation across different reasoning difficulty levels on MMMU. Pruning redundant reasoning steps eliminates logical noise on Easy tasks (+1.8\% gain) and mitigates harmful overthinking on Hard tasks, unlocking a substantial +10.3\% absolute accuracy boost ($\alpha=0.2$) over the Base model.}
  \label{tab:difficulty_accuracy}
  \small
  \begin{tabular}{lcccc}
    \toprule
    \textbf{Difficulty} & \textbf{Baseline} & \textbf{$\alpha=0.1$} & \textbf{$\alpha=0.2$} & \textbf{$\alpha=0.05$} \\
    \midrule
    Easy   & 60.8\% & \textbf{62.6\%} & 62.3\% & 60.5\% \\
    Medium & \textbf{53.2\%} & 47.4\% & 48.2\% & 49.0\% \\
    Hard   & 40.6\% & 44.4\% & \textbf{50.9\%} & 49.1\% \\
    \bottomrule
  \end{tabular}
\end{table}
To further understand the model's reasoning cost under different difficulty levels, 
we evaluated the average number of thinking tokens on the MMMU dataset across the 
Easy, Medium, and Hard categories. As shown in Table~\ref{tab:difficulty_tokens}, 
the results reveal a clear trend: as the reasoning difficulty increases, the model's 
token consumption rises substantially. The Easy category exhibits the lowest token 
usage, the Medium category shows a noticeable increase, and the Hard category requires 
the highest number of tokens, indicating that more complex reasoning tasks demand 
longer reasoning chains and more extensive thinking steps.
At the same time, we observe that the optimized model consistently reduces token 
consumption across all difficulty levels. Compared with the baseline, the reductions 
reach approximately 60\%, 63\%, and 59\% for the Easy, Medium, and Hard categories, 
respectively, demonstrating that the model maintains stable token pruning capability 
regardless of task difficulty.
Taken together, the reasoning cost increases significantly with task complexity, but the 
optimized reasoning strategy achieves stable and effective token compression across 
all difficulty levels, reflecting strong efficiency and controllability in complex 
reasoning scenarios.

To analyze the model's performance under different reasoning difficulty levels, 
we evaluated accuracy on the MMMU dataset across the Easy, Medium, and Hard categories, 
and compared the results under different cross-modal contrastive loss weights 
($\alpha=0.1$, $\alpha=0.2$, $\alpha=0.05$). As shown in Table~\ref{tab:difficulty_accuracy}, 
the baseline model and the various $\alpha$ configurations exhibit distinct accuracy 
patterns across difficulty levels, reflecting how reasoning complexity influences the 
effectiveness of cross-modal alignment.
From the table, we observe that the Easy category achieves its highest accuracy with 
$\alpha=0.1$ (62.6\%), slightly outperforming the baseline. For the Medium category, 
$\alpha=0.05$ yields the best performance (49.0\%), while larger contrastive weights 
lead to noticeable degradation. In the Hard category, $\alpha=0.2$ achieves the highest 
accuracy (50.9\%), significantly surpassing the baseline result of 40.6\%. These findings 
indicate that the model's sensitivity to contrastive loss strength varies with task 
difficulty, and that appropriate alignment strength can enhance performance on more 
challenging reasoning tasks.
These findings illustrate the meaningful role of contrastive loss weight in shaping 
the model's reasoning ability, and highlight that different difficulty levels require 
different alignment strengths to achieve optimal performance.

\section{Conclusion}
In this work, we presented MCPO, a training-efficient framework designed to compress 
multimodal chain-of-thought reasoning while preserving task performance. Through 
comprehensive experiments across MathVista-mini, ScienceQA, MMMU, and MMStar, MCPO 
demonstrated substantial reductions in inference token usage—often exceeding 50\% 
compression—alongside significant end-to-end execution speedups (up to 3.34$\times$), 
while maintaining accuracy within a narrow margin of the baseline model. These findings 
illustrate the effectiveness of combining adaptive step pruning with asymmetric 
modality-contrastive alignment, enabling the model to suppress visual hallucinations 
and retain essential reasoning steps even under aggressive compression.
Our ablation studies further highlight the robustness of MCPO: the method remains 
stable across different contrastive weights and embedding configurations, and 
exhibits consistent behavior across varying reasoning difficulty levels. Together, 
these results validate MCPO as a practical, latency-efficient, and scalable solution 
for multimodal reasoning.
\bibliographystyle{ACM-Reference-Format}
\bibliography{sample-base}

\end{document}